\documentclass[11pt]{article}

\usepackage{acl}

\usepackage{times}
\usepackage{latexsym}

\usepackage[T1]{fontenc}

\usepackage[utf8]{inputenc}

\usepackage{microtype}

\usepackage{inconsolata}

\usepackage{graphicx}

\usepackage{cite}
\usepackage{amsmath,amssymb,amsfonts}

\usepackage{algorithm}
\usepackage{algpseudocode}

\usepackage{graphicx}
\usepackage{textcomp}
\usepackage{xcolor}
\usepackage{tabularx}
\usepackage{varwidth}
\usepackage{multirow}
\usepackage{booktabs}
\usepackage{url}
\usepackage{xcolor} 
\usepackage{booktabs}
\usepackage{array}

\title{Structure-Preserving Document-Level \\ Data Augmentation via Discourse Parsing}

\author{
Ruilin Shang$^{1}$ \quad
Shilong Hou$^{1}$ \quad
Zi Long$^{2,*}$ \quad
Xianghua Fu$^{2}$ \\
$^{1}$Shenzhen University, China \\
$^{2}$Shenzhen Technology University, China \\
\texttt{2410263057@mails.szu.edu.cn} \quad
\texttt{hou\_work@yeah.net} \\
\texttt{longzi@sztu.edu.cn} \quad
\texttt{fuxianghua@sztu.edu.cn} \\
$^{*}$Corresponding author
}

\begin{document}
\maketitle
\begin{abstract}
Data augmentation (DA) plays a vital role in improving model performance in low-resource or limited-supervision scenarios within Natural Language Processing (NLP).  
Existing DA methods, such as synonym replacement and back-translation, have demonstrated effectiveness on the word or sentence level; however, they frequently neglect discourse-level coherence and logical flow, which are essential for complex tasks dependent on inter-sentential relationships. 
In this paper, we propose a structure-preserving document-level data augmentation framework based on large language models (LLMs) and document-level discourse structure parsing.  
Our approach identifies rhetorical relations between sentence pairs and extracts key phrases, which are then replaced with semantically coherent but topically shifted content while preserving the original discourse structure.
Experimental results on text summarization and question answering show that the proposed augmented data provides complementary supervision and improves
combined-data training across the evaluated tasks. Comparisons with a standard RST parser and controlled ablations further support the proposed discourse representation.
\end{abstract}

\begin{figure*}[t]
  \centering
  \includegraphics[scale=0.45]{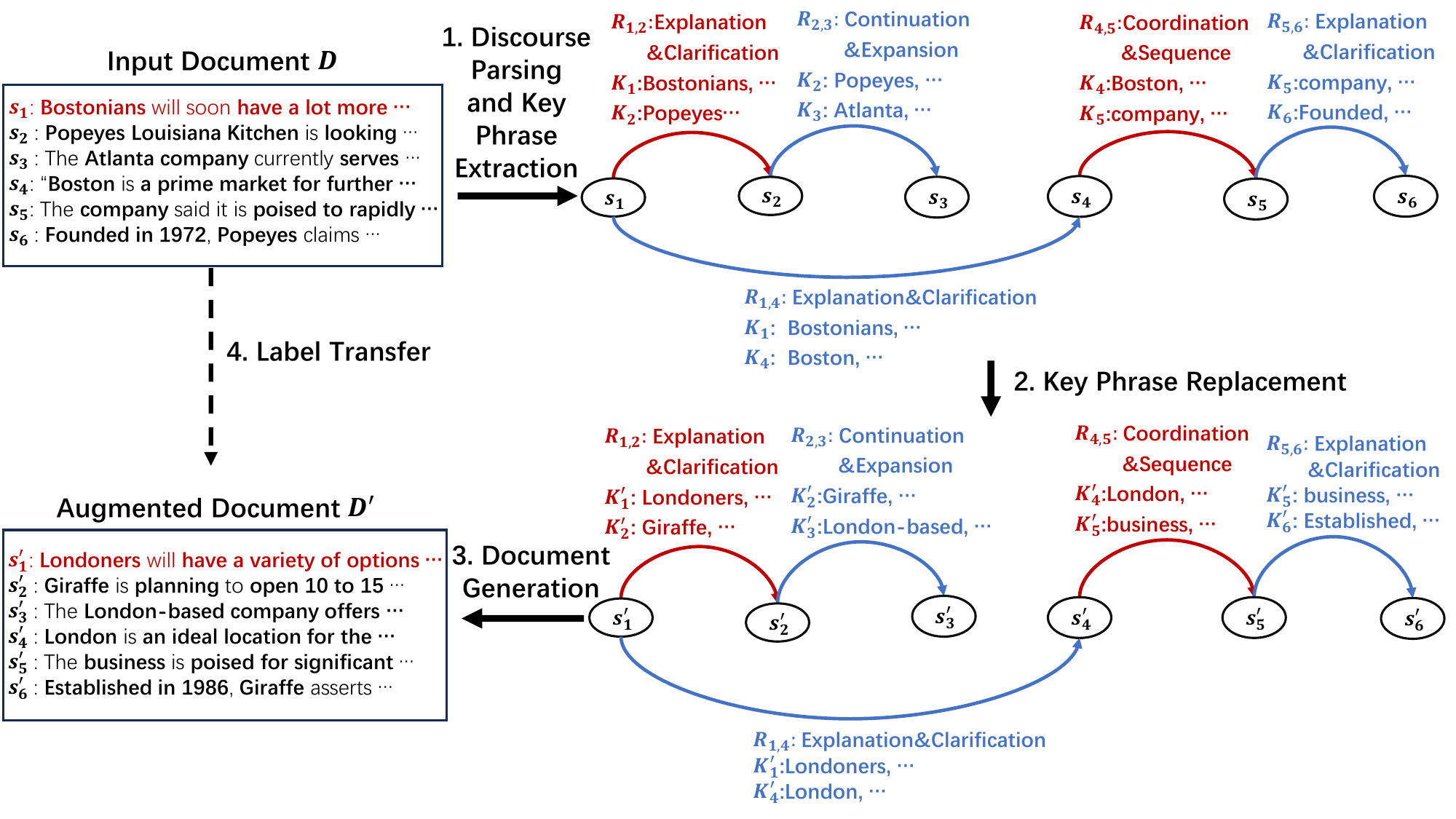}
  \caption{Structure-preserving document-level data augmentation.}
  \label{fig:DA}
\end{figure*}

\section{Introduction}
\label{sec:intro}

Data augmentation (DA) is a pivotal strategy in Natural Language Processing (NLP) for addressing data sparsity in low-resource scenarios by synthetically expanding training sets to improve model robustness and generalization \citep{shorten2019survey,shorten2021text,feng2021survey,chen2023empirical}.
While lexical- and sentence-level DA (e.g., synonym replacement, back-translation) are effective for sentence-level tasks, they may be less effective in document-level contexts like text summarization and question answering (QA). Recent document-level DA methods \citep{huang2023transfer,wu2024importance} frequently overlook higher-level structures, such as inter-sentential coherence and logical flow, leading to semantically inconsistent noise that undermines model training.

To enhance document-level coherence, some researchers incorporate discourse-level information. For instance, \citet{feng2020dialogue} use dialogue-aware strategies for meeting transcripts, \citet{li2025adaptive} apply content reordering, and \citet{jahangir2024complex} generate QA pairs guided by discourse relations. However, these methods usually encode discourse information for a specific task or operation and do not provide a unified mechanism for representing cross-sentence dependencies across tasks and domains.
More importantly, these methods implicitly prioritize content diversity while overlooking document structure and task-level supervision.
We further argue that, in document-level data augmentation, an important challenge is label invalidation caused by structural distortion.

Preserving discourse structure during augmentation retains the sentence-level units on which downstream labels are defined. For summarization, labels can be
transferred through sentence indices. For extractive QA, the transformed answer can be located within the aligned sentence to reconstruct its new token span. This avoids regenerating task labels with an LLM, which may introduce additional noise or bias. Structural preservation therefore supports label reuse while allowing the document content to change.

We propose a structure-preserving DA framework based on document-level discourse structure analysis. Our method models rhetorical relations and key phrases between sentence pairs, replacing content while preserving sentence order, count, and discourse structure. This ensures sentence-level alignment between original and augmented documents, enabling consistent label transfer. Experiments on summarization and QA tasks demonstrate the effectiveness of the proposed augmentation across multiple datasets and training settings.

Our contributions are summarized as follows:
\begin{itemize}
\setlength{\itemsep}{0pt}
\setlength{\parsep}{0pt}
    \item A document-level discourse structure analysis framework that structurally models rhetorical relations and sentence-level key semantics.
    \item A novel discourse-level DA method that manipulates core content while preserving rhetorical structures to generate coherent, diverse instances.
    \item Extensive experiments across four languages, two tasks, multiple model architectures, and controlled discourse variants demonstrate the robustness of the proposed framework.
\end{itemize}

\section{Related Work}
\label{sec:relate}

Data augmentation (DA) expands training data through operations based on prior knowledge \citep{wei2019eda} for supervised \citep{wei2019eda} or semi-supervised learning \citep{xie2020unsupervised}. Current methods span token \citep{niu2018adversarial,kumar2020data}, sentence \citep{sennrich2016improving,chen2020semi}, adversarial \citep{cheng2019robust,morris2020textattack}, and hidden-space \citep{malandrakis2019controlled,chen2021hiddencut} levels. Recently, LLMs \citep{ye2022zerogen,chowdhury2024generative,sahu2025guide} have significantly advanced DA with their generative power.

While effective at the sentence level, standard DA may disrupt coherence in document-level tasks \citep{huang2023transfer,wu2024importance} because it
does not explicitly preserve inter-sentential dependencies. Recent studies therefore incorporate discourse information into data augmentation
\citep{pasunuru2021data,jahangir2024complex,
li2025adaptive}. These approaches, however, are generally designed for particular tasks or use selected discourse cues without constructing a reusable
document-level representation. In contrast, our document-level discourse structure analysis framework jointly represents antecedent links, rhetorical relations, and sentence-level key semantics, and uses this structure to constrain document generation and label transfer.

Regarding discourse parsing, foundational frameworks like RST \citep{mann1988rhetorical} and PDTB \citep{prasad2008penn} provide structured representations via rhetorical trees or argument spans, with recent advances driven by neural parsers \citep{morey2017much,shi2020end,hu2023rst}. 
Recent unified RST parsers support multiple languages and relation inventories \citep{chistova-2025-bridging}. We adopt a simpler sentence-level representation designed for aligned generation and compare it with UniRST in Section~\ref{sec:discourse}.

\begin{algorithm}[t]
\caption{Structure-Preserving Document-Level Data Augmentation}
\label{alg:sp-augmentation}
\footnotesize
\begin{algorithmic}[1]
\Require Document $\mathbf{D}=\{s_1,\ldots,s_n\}$
\Ensure Augmented document $\mathbf{D}'=\{s'_1,\ldots,s'_n\}$

\Statex
\Statex \textbf{Stage 1: Discourse parsing and key phrase extraction}
\State $\mathrm{OriList} \gets [\,]$

\For{$j \gets 2$ \textbf{to} $n$}
    \State $s_i \gets
    \Call{FindValid}{s_1,\ldots,s_{j-1}}$
    \State $R_{i,j} \gets \alpha(s_i,s_j)$
    \State $\mathbf{K}_i \gets \beta(s_i)$
    \Comment{$\mathbf{K}_i=(k_i^1,\ldots,k_i^p)$}
    \State $\mathbf{K}_j \gets \beta(s_j)$
    \Comment{$\mathbf{K}_j=(k_j^1,\ldots,k_j^q)$}
    \State $\mathbf{A}_{i,j} \gets
    (s_i,\mathbf{K}_i,s_j,\mathbf{K}_j,R_{i,j})$
    \State $\mathrm{OriList}.\Call{append}{\mathbf{A}_{i,j}}$
\EndFor

\Statex
\Statex \textbf{Stage 2: Key phrase replacement}
\State $\mathrm{AugList} \gets [\,]$

\ForAll{$\mathbf{A}_{i,j} \in \mathrm{OriList}$}
    \State $\mathbf{K}'_{<i} \gets
    (\mathbf{K}'_1,\ldots,\mathbf{K}'_{i-1})$
    \State $\mathbf{K}'_i \gets
    \gamma(\mathbf{K}_i,\mathbf{K}'_{<i})$
    \Comment{$\mathbf{K}'_i=(k_i^{\prime 1},\ldots,k_i^{\prime p})$}

    \State $\mathbf{K}'_{<j} \gets
    (\mathbf{K}'_1,\ldots,\mathbf{K}'_{j-1})$
    \State $\mathbf{K}'_j \gets
    \gamma(\mathbf{K}_j,\mathbf{K}'_{<j})$
    \Comment{$\mathbf{K}'_j=(k_j^{\prime 1},\ldots,k_j^{\prime q})$}

    \State $\mathbf{A}'_{i,j} \gets
    (s_i,\mathbf{K}'_i,s_j,\mathbf{K}'_j,R_{i,j})$
    \State $\mathrm{AugList}.\Call{append}{\mathbf{A}'_{i,j}}$
\EndFor

\Statex
\Statex \textbf{Stage 3: Document generation}
\State $\mathbf{D}' \gets [\,]$

\For{$i \gets 1$ \textbf{to} $n$}
    \State $s'_i \gets
    \delta(\mathbf{A}'_{i,*},\mathbf{A}'_{*,i})$
    \State $\mathbf{D}'.\Call{append}{s'_i}$
\EndFor

\State \Return $\mathbf{D}'$
\end{algorithmic}
\end{algorithm}

\section{Methodology}
\label{sec:method}

As shown in Algorithm~\ref{alg:sp-augmentation} and Fig.~\ref{fig:DA}, we propose a structure-preserving data augmentation framework defined by three invariants: (1) unchanged document structure, (2) preserved rhetorical relations, and (3) identical sentence count and ordering. These ensure that sentence-level supervision remains aligned, i.e., the $i$-th sentence in the augmented document $\mathbf{D}'$ corresponds exactly to the $i$-th sentence in the original $\mathbf{D}$.

\subsection{Discourse Structure Analysis and Key Phrase Extraction}
\label{sec:parsing}

Given $\mathbf{D} = {s_{1}, \ldots, s_{n}}$, we identify rhetorical structures and key semantics (Algorithm~\ref{alg:sp-augmentation}, Lines 1--9), adopting a simplified alternative to standard RST tailored
to sentence-aligned generation. Each sentence $s_j$ ($j \ge 2$) is linked to a single antecedent $s_i$ ($i < j$) via a relation $R_{i,j}$.
We define six relation categories: Causality \& Condition, Contrast \& Transition, Coordination \& Sequence, Explanation \& Clarification, Summary \& Generalization, and Continuation \& Expansion. This design balances expressiveness and robustness, covering common inter-sentential functions while reducing the sparsity associated with fine-grained taxonomies.
\footnote{As a small-scale sanity check, we manually annotated
100 sentence pairs sampled from Newsroom. After
mapping both systems to the same six-relation inventory,
our predictions matched the manual annotations in
78 cases, compared with 41 for DPLP \citep{ji2014representation}.
Given the limited sample size, we treat this result as
supporting evidence rather than a complete
discourse-parsing benchmark.}

The function \textsc{FindValid} first examines the immediately preceding sentence $s_{j-1}$. If no valid relation is found, it scans $s_1,\ldots,s_{j-2}$. When the scan also fails, $s_j$ is linked to $s_{j-1}$ using Continuation \& Expansion as the default relation. This adjacent-first heuristic does not change the $O(n^2)$ worst-case complexity, but it reduces the number of relation checks when a local antecedent is accepted. Function $\alpha(s_i, s_j)$ predicts the rhetorical relation $R_{i,j}$ between two sentences. Function $\beta(s_i)$ extracts salient semantic phrases $\mathbf{K}i$ that capture the core content of a sentence, such as named entities, noun phrases, event triggers, and domain-specific concepts. The extracted information is represented as $\mathbf{A}{i,j} = (s_i, \mathbf{K}_i, s_j, \mathbf{K}j, R{i,j})$.

\subsection{Key Phrase Replacement}
\label{sec:replacement}

In Lines 10--18, we generate augmented key phrases $\mathbf{K}'_i, \mathbf{K}'_j$ while preserving $R_{i,j}$. 
The function $\gamma$ replaces each $k \in \mathbf{K}$ with a semantically consistent but topically distinct alternative, referencing previously replaced phrases $(\mathbf{K}'_{1},\cdots, \mathbf{K}'_{i-1})$ to maintain thematic consistency. 
We enforce a globally consistent mapping such that identical source key phrases are consistently replaced with the same target phrase. To further constrain semantic drift, a target topic is randomly sampled in advance, and all replacements are conditioned on this topic rather than generated freely. This results in structure-preserved augmented placeholders $\mathbf{A}'_{i,j} = (s_{i}, \mathbf{K}'_i, s_{j}, \mathbf{K}'_j, R_{i,j})$.

\subsection{Document Generation}
\label{sec:generation}

Finally, $\mathbf{D}'$ is generated by assembling sentences $\{s'_1, \ldots, s'_n\}$ (Lines 19--24). For each sentence index $i$, function
$\delta(A'_{i,*}, A'_{*,i})$ synthesizes a new sentence
$s'_i$ that incorporates the modified key phrases while
adhering to the original rhetorical relations $R_{i,j}$. This stage maintains global structural consistency while introducing semantic diversity.

\subsection{Label Transfer}
\label{sec:label}

As the document layout is preserved, target-side annotations can be derived via rule-based mapping. For text summarization, if a sentence $s_i$ in the original document is selected as part of the summary, its augmented counterpart $s'_i$ is assigned the same label, ensuring consistent supervision under sentence-level alignment.

For question answering, we perform token-level label transfer via span reconstruction. During key phrase replacement, the same transformation $\gamma$ is applied to the context, question, and answer to maintain consistency.
Instead of reusing original token spans, we locate the corresponding transformed answer string in the augmented context to reconstruct the answer span. As sentence order is preserved, the answer remains within the corresponding sentence $s_i'$ aligned with the original $s_i$, and key-phrase-level replacements ensure a consistent mapping between original and transformed answer tokens.

\begin{table*}[t]
\centering
\small
\setlength{\tabcolsep}{2.65pt}
\renewcommand{\arraystretch}{0.96}

\begin{tabular}{@{}lcccccccccccc@{}}
\toprule
\multirow{2}{*}{\textbf{Training Configuration}}
& \multicolumn{4}{c}{\textbf{Chinese Abstractive Corpus}}
& \multicolumn{4}{c}{\textbf{Newsroom}}
& \multicolumn{4}{c}{\textbf{IndoSum}} \\
\cmidrule(lr){2-5}
\cmidrule(lr){6-9}
\cmidrule(lr){10-13}
& \textbf{B}
& \textbf{R1}
& \textbf{R2}
& \textbf{RL}
& \textbf{B}
& \textbf{R1}
& \textbf{R2}
& \textbf{RL}
& \textbf{B}
& \textbf{R1}
& \textbf{R2}
& \textbf{RL} \\
\midrule

\multicolumn{13}{@{}l}{\textit{Reference Settings}} \\
\addlinespace[1pt]

Zero-shot LLM
\textsuperscript{\ref{fn:qwen3b}}
& 0.139 & 0.290 & 0.193 & 0.265
& 0.267 & 0.401 & 0.290 & 0.364
& 0.151 & 0.290 & 0.248 & 0.273 \\

Inp
& 0.249 & 0.391 & 0.312 & 0.372
& 0.634 & 0.697 & 0.657 & 0.686
& 0.728 & 0.749 & 0.741 & 0.747 \\

Con
& 0.228 & 0.370 & 0.289 & 0.351
& 0.653 & 0.712 & 0.676 & 0.703
& 0.746 & 0.765 & 0.756 & 0.763 \\

Inp+Con
& 0.269 & 0.392 & 0.318 & 0.376
& 0.686 & 0.741 & 0.709 & 0.733
& 0.757 & 0.783 & 0.776 & 0.782 \\

\midrule

\multicolumn{13}{@{}l}{
\textit{Back-Translation} \citep{sennrich2016improving}
} \\
\addlinespace[1pt]

Aug
& 0.211 & 0.310 & 0.247 & 0.298
& 0.752 & 0.769 & 0.739 & 0.761
& 0.498 & 0.516 & 0.466 & 0.505 \\

Inp+Aug
& 0.210 & 0.337 & 0.259 & 0.319
& 0.768 & 0.775 & 0.744 & 0.766
& 0.601 & 0.663 & 0.577 & 0.643 \\

Inp+Con+Aug
& 0.198 & 0.335 & 0.253 & 0.315
& 0.774 & 0.759 & 0.730 & 0.751
& 0.672 & 0.689 & 0.647 & 0.671 \\

\midrule

\multicolumn{13}{@{}l}{
\textit{MixSumm} \citep{sahu2025guide}
} \\
\addlinespace[1pt]

Aug
& \textbf{0.296} & 0.393
& \underline{0.329} & 0.378
& 0.147 & 0.293 & 0.161 & 0.250
& 0.358 & 0.415 & 0.389 & 0.407 \\

Inp+Aug
& 0.223 & 0.344 & 0.263 & 0.325
& 0.463 & 0.545 & 0.462 & 0.518
& 0.582 & 0.637 & 0.629 & 0.631 \\

Inp+Con+Aug
& 0.193 & 0.317 & 0.236 & 0.297
& 0.611 & 0.624 & 0.570 & 0.608
& 0.560 & 0.615 & 0.588 & 0.609 \\

\midrule

\multicolumn{13}{@{}l}{
\textit{LLM-based Summary Relabeling}
\textsuperscript{\ref{fn:qwen7b}}
} \\
\addlinespace[1pt]

Aug
& 0.221 & 0.345 & 0.271 & 0.329
& 0.599 & 0.656 & 0.597 & 0.638
& 0.380 & 0.442 & 0.421 & 0.435 \\

Inp+Aug
& 0.186 & 0.329 & 0.243 & 0.309
& 0.748 & 0.756 & 0.722 & 0.747
& 0.572 & 0.592 & 0.517 & 0.571 \\

Inp+Con+Aug
& 0.165 & 0.312 & 0.223 & 0.291
& 0.751 & 0.767 & 0.707 & 0.730
& 0.590 & 0.645 & 0.579 & 0.625 \\

\midrule

\multicolumn{13}{@{}l}{
\textit{Direct LLM Augmentation}
\textsuperscript{\ref{fn:qwen7b}}
} \\
\addlinespace[1pt]

Aug
& 0.198 & 0.344 & 0.256 & 0.324
& 0.338 & 0.446 & 0.337 & 0.406
& 0.317 & 0.376 & 0.306 & 0.375 \\

Inp+Aug
& 0.193 & 0.340 & 0.252 & 0.320
& 0.590 & 0.645 & 0.588 & 0.626
& 0.401 & 0.463 & 0.397 & 0.461 \\

Inp+Con+Aug
& 0.214 & 0.354 & 0.270 & 0.334
& 0.677 & 0.724 & 0.676 & 0.708
& 0.598 & 0.615 & 0.589 & 0.607 \\

\midrule

\multicolumn{13}{@{}l}{
\textit{Our Method}
} \\
\addlinespace[1pt]

Aug
& 0.247 & 0.383 & 0.304 & 0.360
& 0.639 & 0.696 & 0.658 & 0.686
& 0.703 & 0.758 & 0.751 & 0.757 \\

Inp+Aug
& 0.269 & \underline{0.395}
& 0.319 & \underline{0.379}
& \underline{0.775} & \textbf{0.796}
& \textbf{0.765} & \textbf{0.787}
& \underline{0.779} & \underline{0.792}
& \underline{0.786} & \underline{0.789} \\

Inp+Con+Aug
& \underline{0.276} & \textbf{0.417}
& \textbf{0.343} & \textbf{0.400}
& \textbf{0.780} & \underline{0.789}
& \underline{0.757} & \underline{0.780}
& \textbf{0.782} & \textbf{0.801}
& \textbf{0.795} & \textbf{0.799} \\

\bottomrule
\end{tabular}

\vspace{2pt}

\caption{
Performance of fine-tuning and data augmentation methods
on the Chinese Abstractive Corpus, Newsroom, and IndoSum.
B, R1, R2, and RL denote BLEU, ROUGE-1, ROUGE-2,
and ROUGE-L.
}
\label{tab:summary}
\end{table*}

\section{Experiment Settings}
\label{sec:exp}

\subsection{Datasets and Metrics}
\label{sec:dataset}

We evaluate the framework on text summarization and extractive QA. For summarization, we use the Chinese Abstractive Corpus\footnote{\url{https://github.com/wonderfulsuccess/chinese_abstractive_corpus}}, Newsroom \citep{N18-1065}, and IndoSum \citep{kurniawan2018indosum}. For QA, we use SQuAD 1.1 \citep{rajpurkar2016squad} and the Japanese dataset JaQuAD \citep{so2022jaquad}.

For summarization, we formulate supervision at the
sentence level. Reference summaries are aligned to source sentences by exact sentence matching during preprocessing, and the resulting sentence indices are used as supervision. The same indices are then assigned
to the aligned sentences in the augmented document. To support sentence-level label transfer on IndoSum, we retain documents whose reference summary forms a contiguous span in the source document.

For each summarization dataset, we sample 2,000
instances for the input set, 2,000 for the control set, and 1,000 for testing.
For each QA dataset, we use 4,000 input instances,
4,000 control instances, and 1,000 test instances.
Within each dataset, input, control, and the test set are mutually exclusive, and the augmented set is generated only from the input set.

Summarization performance is measured by BLEU \citep{papineni2002bleu}, ROUGE-1, ROUGE-2, and ROUGE-L \citep{lin2004rouge}. 
QA performance is evaluated using F1 score and Exact Match (EM) \citep{rajpurkar2016squad}.

\subsection{Training Configurations and Baselines}
\label{sec:baseline}

To systematically evaluate the proposed structure-preserving DA approach, we fine-tune a base LLM under six distinct training configurations:
(1) \textbf{Inp}, which uses only the sampled input data;
(2) \textbf{Con}, which uses only the control data;
(3) \textbf{Aug}, which uses only the augmented data generated from Inp  by the augmentation method being evaluated;
(4) \textbf{Inp+Con}, which combines the input and control data;
(5) \textbf{Inp+Aug}, which combines the input and augmented data; and
(6) \textbf{Inp+Con+Aug}, which integrates the input, control, and augmented data.
For our method, each augmented document maintains one-to-one alignment with the original in sentence order and count, enabling direct transfer of sentence-level supervision.

We compare with several DA strategies. For both tasks,
we include Back-Translation \citep{sennrich2016improving}, which modifies only
the source document or QA context while retaining the
original supervision; in QA, the question and answer
remain unchanged. For summarization, we additionally
include MixSumm \citep{sahu2025guide}, while for QA
we include Two-Step QA Augmentation \citep{chowdhury2024generative}.

To assess label reusability, we include LLM-based
Summary Relabeling and LLM-based QA Relabeling,
which replace transferred labels with labels regenerated
by Qwen2.5-7B-Instruct \citep{qwen2025qwen25technicalreport}\footnote{\url{https://huggingface.co/Qwen/Qwen2.5-7B-Instruct} \label{fn:qwen7b}}.
Direct LLM Augmentation instead generates augmented
instances and labels without our discourse constraints.
All LLM-based augmentation baselines use
Qwen2.5-7B-Instruct.

In Tables~\ref{tab:summary} and~\ref{tab:QA}, Aug, Inp+Aug, and Inp+Con+Aug denote the corresponding training configurations under each augmentation method.

\subsection{Implementation Details}
\label{sec:impl}

For data augmentation, we use
Qwen2.5-7B-Instruct
to perform discourse structure analysis, key-phrase
replacement, and document generation. The augmentation
process is decomposed into three independently prompted
stages, each executed with a single prompt.

For downstream tasks,
Qwen2.5-3B-Instruct\footnote{\label{fn:qwen3b}\url{https://huggingface.co/Qwen/Qwen2.5-3B-Instruct}}
is used for zero-shot evaluation. We fine-tune
Qwen2.5-3B-Base\footnote{\url{https://huggingface.co/Qwen/Qwen2.5-3B}}
using LoRA \citep{hu2022lora} with rank 8. The batch size is 1 per device with gradient accumulation of 32. We train for 3 epochs with a learning rate of $1 \times 10^{-5}$, using a cosine scheduler with 0.1 warmup ratio. All experiments are conducted in bfloat16 precision.

\begin{table}[t]
\centering
\small
\setlength{\tabcolsep}{2.5pt}
\renewcommand{\arraystretch}{0.95}

\begin{tabularx}{\columnwidth}{
@{}
>{\raggedright\arraybackslash}X
cc@{\hspace{4pt}}cc
@{}
}
\toprule
\textbf{Training Configuration}
& \multicolumn{2}{c}{\textbf{SQuAD}}
& \multicolumn{2}{c}{\textbf{JaQuAD}} \\
\cmidrule(lr){2-3}
\cmidrule(lr){4-5}
& \textbf{EM} & \textbf{F1}
& \textbf{EM} & \textbf{F1} \\
\midrule

\textit{Reference Settings}
& & & & \\[1pt]

Zero-shot LLM\textsuperscript{\ref{fn:qwen3b}}
& 0.636
& \textbf{0.791}
& 0.659
& 0.773 \\

Inp
& \underline{0.708}
& \underline{0.780}
& 0.547
& 0.713 \\

Con
& 0.690
& 0.742
& 0.598
& 0.743 \\

Inp+Con
& 0.697
& 0.746
& \underline{0.681}
& 0.729 \\

\midrule

\textit{Back-Translation} \\
\citep{sennrich2016improving}
& & & & \\[1pt]

Aug
& 0.362
& 0.508
& 0.117
& 0.433 \\

Inp+Aug
& 0.426
& 0.559
& 0.381
& 0.585 \\

Inp+Con+Aug
& 0.514
& 0.640
& 0.548
& 0.774 \\

\midrule

\textit{Two-Step QA Augmentation}
\citep{chowdhury2024generative}
& & & & \\[1pt]

Aug
& 0.276
& 0.479
& 0.166
& 0.489 \\

Inp+Aug
& 0.470
& 0.622
& 0.503
& 0.735 \\

Inp+Con+Aug
& 0.421
& 0.542
& 0.531
& 0.759 \\

\midrule

\textit{LLM-based QA Relabeling}
\textsuperscript{\ref{fn:qwen7b}}
& & & & \\[1pt]

Aug
& 0.315
& 0.460
& 0.209
& 0.500 \\

Inp+Aug
& 0.370
& 0.515
& 0.529
& 0.741 \\

Inp+Con+Aug
& 0.577
& 0.606
& 0.595
& \underline{0.787} \\

\midrule

\textit{Direct LLM Augmentation}
\textsuperscript{\ref{fn:qwen7b}}
& & & & \\[1pt]

Aug
& 0.302
& 0.444
& 0.110
& 0.433 \\

Inp+Aug
& 0.313
& 0.450
& 0.442
& 0.681 \\

Inp+Con+Aug
& 0.576
& 0.677
& 0.524
& 0.780 \\

\midrule

\textit{Our Method}
& & & & \\[1pt]

Aug
& 0.698
& 0.740
& 0.546
& 0.728 \\

Inp+Aug
& 0.701
& 0.751
& \textbf{0.688}
& 0.771 \\

Inp+Con+Aug
& \textbf{0.731}
& 0.779
& 0.658
& \textbf{0.811} \\

\bottomrule
\end{tabularx}

\caption{
Performance of fine-tuning and data augmentation
methods on the English SQuAD and Japanese JaQuAD
question answering tasks.
}
\label{tab:QA}
\end{table}

\section{Results}
\label{sec:res_ana}

\subsection{Results on Text Summarization}
\label{sec:sum}

\paragraph{Utility of augmented data.}

Table~\ref{tab:summary} reports the results on three
summarization datasets. On the Chinese corpus,
Inp+Con+Aug improves over Inp+Con from 0.269 to
0.276 in BLEU and from 0.376 to 0.400 in ROUGE-L.
Although MixSumm obtains the highest standalone BLEU score under the Aug configuration, its performance decreases when mixed with real data. In contrast, the proposed augmentation remains
effective in the combined-data setting and achieves the
best ROUGE-1, ROUGE-2, and ROUGE-L scores.

The complementary effect is stronger on Newsroom.
Adding Aug to Inp increases BLEU from 0.634 to 0.775,
while adding Aug to Inp+Con increases it from 0.686 to
0.780. Inp+Aug achieves the best ROUGE scores, and
Inp+Con+Aug achieves the best BLEU score. On
IndoSum, Inp+Con+Aug obtains the best result on all four
metrics, improving BLEU from 0.757 to 0.782 and
ROUGE-L from 0.782 to 0.799 over Inp+Con. These
results show that the generated documents provide
complementary training signals rather than merely
duplicating the input set.

\paragraph{Reliability of transferred labels.}
The comparison with LLM-based Summary Relabeling separates the quality of augmented documents from the reliability of their supervision. Under Inp+Aug, the proposed method achieves BLEU scores of 0.269, 0.775, and 0.779 on the Chinese corpus, Newsroom, and IndoSum, respectively. Under the same Inp+Aug configuration, LLM-based Summary Relabeling obtains 0.186, 0.748, and 0.572. The gaps become particularly large on the Chinese corpus and IndoSum.

These results indicate that regenerating summary labels
introduces additional supervision noise. Our framework instead transfers the original sentence indices to the aligned augmented sentences, avoiding
additional label regeneration.
Back-translation remains competitive on Newsroom, but
its advantage is smaller in combined-data settings.
Overall, the results suggest that both content diversity
and label compatibility contribute to effective
document-level augmentation.

\begin{table*}[t]
\centering
\small
\setlength{\tabcolsep}{2.7pt}
\renewcommand{\arraystretch}{0.96}

\begin{tabular}{llcccccccccc}
\toprule
\multirow{2}{*}{\textbf{Method}}
& \multirow{2}{*}{\textbf{Configuration}}
& \multicolumn{4}{c}{\textbf{Chinese Abstractive Corpus}}
& \multicolumn{4}{c}{\textbf{Newsroom}}
& \multicolumn{2}{c}{\textbf{SQuAD}} \\
\cmidrule(lr){3-6}
\cmidrule(lr){7-10}
\cmidrule(lr){11-12}
& 
& \textbf{BLEU}
& \textbf{R1}
& \textbf{R2}
& \textbf{RL}
& \textbf{BLEU}
& \textbf{R1}
& \textbf{R2}
& \textbf{RL}
& \textbf{EM}
& \textbf{F1} \\
\midrule

\multicolumn{12}{l}{\textit{Proposed method}} \\[1pt]

\multirow{3}{*}{Full method}
& Aug
& 0.247 & 0.383 & 0.304 & 0.360
& 0.639 & 0.696 & 0.658 & 0.686
& 0.698 & 0.740 \\

& Inp+Aug
& \underline{0.269}
& \underline{0.395}
& 0.319
& \underline{0.379}
& \underline{0.775}
& \textbf{0.796}
& \textbf{0.765}
& \textbf{0.787}
& \underline{0.701}
& \underline{0.751} \\

& Inp+Con+Aug
& \textbf{0.276}
& \textbf{0.417}
& \textbf{0.343}
& \textbf{0.400}
& \textbf{0.780}
& \underline{0.789}
& \underline{0.757}
& \underline{0.780}
& \textbf{0.731}
& \textbf{0.779} \\

\midrule
\multicolumn{12}{l}{\textit{Comparison with UniRST}} \\[1pt]

\multirow{3}{*}{UniRST}
& Aug
& 0.164 & 0.303 & 0.216 & 0.283
& 0.285 & 0.387 & 0.268 & 0.344
& 0.393 & 0.559 \\

& Inp+Aug
& 0.184 & 0.314 & 0.236 & 0.298
& 0.580 & 0.639 & 0.582 & 0.620
& 0.490 & 0.629 \\

& Inp+Con+Aug
& 0.174 & 0.299 & 0.220 & 0.282
& 0.731 & 0.768 & 0.731 & 0.755
& 0.506 & 0.692 \\

\midrule
\multicolumn{12}{l}{\textit{Ablation study}} \\[2pt]

\multirow{3}{*}{w/o Relation}
& Aug
& 0.189 & 0.230 & 0.199 & 0.222
& 0.356 & 0.522 & 0.477 & 0.510
& 0.455 & 0.582 \\

& Inp+Aug
& 0.208 & 0.338 & 0.285 & 0.320
& 0.448 & 0.559 & 0.529 & 0.552
& 0.515 & 0.606 \\

& Inp+Con+Aug
& 0.240 & 0.381 & 0.326 & 0.362
& 0.618 & 0.639 & 0.606 & 0.630
& 0.570 & 0.615 \\

\addlinespace[1pt]

\multirow{3}{*}{Adjacent-only}
& Aug
& 0.194 & 0.252 & 0.218 & 0.243
& 0.348 & 0.558 & 0.527 & 0.552
& 0.384 & 0.395 \\

& Inp+Aug
& 0.226 & 0.347 & 0.302 & 0.334
& 0.423 & 0.581 & 0.556 & 0.575
& 0.459 & 0.483 \\

& Inp+Con+Aug
& 0.240 & 0.383
& \underline{0.328}
& 0.364
& 0.654 & 0.614 & 0.591 & 0.609
& 0.573 & 0.591 \\

\bottomrule
\end{tabular}

\caption{
Comparison with UniRST and controlled discourse
ablations under matched training configurations.
}
\label{tab:discourse-analysis}
\end{table*}

\begin{table*}[t]
\centering

\small
\setlength{\tabcolsep}{3.1pt}
\renewcommand{\arraystretch}{1.00}

\begin{tabular}{@{}llcccccccc@{}}
\toprule
\multirow{2}{*}{\textbf{Dataset}}
& \multirow{2}{*}{\textbf{\shortstack{Training\\Configuration}}}
& \multicolumn{4}{c}{\textbf{BertSum}}
& \multicolumn{4}{c}{\textbf{SimCLS}} \\
\cmidrule(lr){3-6}
\cmidrule(lr){7-10}
&
& \textbf{BLEU}
& \textbf{R-1}
& \textbf{R-2}
& \textbf{R-L}
& \textbf{BLEU}
& \textbf{R-1}
& \textbf{R-2}
& \textbf{R-L} \\
\midrule

\multirow{9}{*}{\shortstack[l]{Chinese Abstractive\\Corpus}}
& Inp
& \underline{0.430}
& \underline{0.515}
& \underline{0.462}
& \underline{0.502}
& \underline{0.439}
& \underline{0.522}
& \underline{0.466}
& \underline{0.507} \\

& Con
& 0.411 & 0.497 & 0.440 & 0.482
& 0.423 & 0.507 & 0.450 & 0.492 \\

& Aug
& 0.415 & 0.496 & 0.445 & 0.484
& \underline{0.439} & 0.517 & 0.463 & 0.503 \\

& Inp+Con
& 0.422 & 0.505 & 0.452 & 0.492
& 0.424 & 0.509 & 0.453 & 0.493 \\

& Inp+Aug
& 0.422 & 0.508 & 0.451 & 0.493
& 0.431 & 0.516 & 0.458 & 0.501 \\

& Inp+Con+Aug
& \textbf{0.442}
& \textbf{0.526}
& \textbf{0.471}
& \textbf{0.512}
& \textbf{0.442}
& \textbf{0.526}
& \textbf{0.470}
& \textbf{0.511} \\

& Aug-MixSumm
& 0.237 & 0.351 & 0.267 & 0.328
& 0.354 & 0.443 & 0.375 & 0.424 \\

& Inp+Aug-MixSumm
& 0.408 & 0.496 & 0.435 & 0.480
& 0.418 & 0.503 & 0.445 & 0.488 \\

& Inp+Con+Aug-MixSumm
& 0.303 & 0.402 & 0.335 & 0.387
& 0.425 & 0.509 & 0.451 & 0.493 \\

\midrule

\multirow{9}{*}{Newsroom}
& Inp
& 0.857 & 0.861 & 0.840 & 0.855
& \underline{0.858} & 0.860 & 0.840 & 0.854 \\

& Con
& 0.850 & 0.857 & 0.834 & 0.851
& 0.840 & 0.846 & 0.822 & 0.839 \\

& Aug
& 0.851 & 0.857 & 0.836 & 0.852
& 0.856 & 0.862 & 0.841 & 0.856 \\

& Inp+Con
& 0.854 & 0.856 & 0.836 & 0.851
& 0.852 & 0.860 & 0.837 & 0.850 \\

& Inp+Aug
& \underline{0.862}
& \underline{0.864}
& \underline{0.845}
& \underline{0.860}
& \textbf{0.877}
& \underline{0.875}
& \underline{0.858}
& \underline{0.871} \\

& Inp+Con+Aug
& \textbf{0.870}
& \textbf{0.872}
& \textbf{0.855}
& \textbf{0.867}
& \textbf{0.877}
& \textbf{0.879}
& \textbf{0.861}
& \textbf{0.874} \\

& Aug-MixSumm
& 0.142 & 0.292 & 0.153 & 0.244
& 0.271 & 0.395 & 0.270 & 0.349 \\

& Inp+Aug-MixSumm
& 0.589 & 0.644 & 0.580 & 0.623
& 0.783 & 0.798 & 0.765 & 0.787 \\

& Inp+Con+Aug-MixSumm
& 0.808 & 0.820 & 0.792 & 0.812
& 0.827 & 0.837 & 0.810 & 0.829 \\

\bottomrule
\end{tabular}
\caption{Performance of BertSum and SimCLS on the text summarization task. Inp, Con, and Aug denote the original input, control data, and our augmented data, respectively. Aug-MixSumm denotes the training configuration using MixSumm-generated augmented data ~\citep{sahu2025guide}.}
\label{tab:summary-2}
\end{table*}

\subsection{Question Answering Results}
\label{sec:qa}

\paragraph{Results on SQuAD.}
Table~\ref{tab:QA} reports the results on SQuAD and
JaQuAD. On SQuAD, Aug alone reaches 0.698 EM, close
to the 0.708 obtained by Inp, indicating that the
augmented contexts retain substantial answer supervision.
Adding Aug to Inp+Con improves EM from 0.697 to
0.731 and F1 from 0.746 to 0.779. The resulting F1 is
also close to the best fine-tuned result of 0.780 obtained
by Inp. Thus, the main improvement is not limited to
token overlap but also appears in exact answer matching.

The comparison with regenerated labels provides further evidence for the role of alignment. Under Inp+Aug, our method obtains 0.701 EM and 0.751 F1, whereas LLM-based QA Relabeling obtains only 0.370 and 0.515. Back-Translation, Two-Step QA Augmentation, and Direct LLM Augmentation are also weaker under the corresponding configurations. These methods may alter the relationship among the context, question, and answer, while our method applies consistent transformations and reconstructs the answer span.

\paragraph{Results on JaQuAD.}
JaQuAD shows a different balance between EM and F1.
Inp+Aug obtains the highest EM score of 0.688, whereas
Inp+Con+Aug obtains the highest F1 score of 0.811.
Compared with Inp+Con, adding Aug increases F1 from
0.729 to 0.811, although EM decreases from 0.681 to
0.658. This divergence suggests that augmented training
improves answer-token coverage but does not always
produce the exact reference boundary or surface form.

Nevertheless, under Inp+Con+Aug, our method obtains
higher numerical EM and F1 scores than the corresponding
augmentation baselines. The results therefore support cross-lingual applicability while also showing
that exact matching remains sensitive to language-specific answer realization.

\section{Analysis}

\subsection{Analysis of Discourse Structure}
\label{sec:discourse}

\paragraph{Multilingual relation validation.}
To examine whether the six-relation inventory transfers
across languages, we manually annotate 100 sentence pairs
from each of the Chinese, English, Indonesian, and
Japanese datasets. The predicted relations achieve
accuracies of 81\%, 78\%, 71\%, and 74\%, respectively,
supporting the cross-lingual applicability of the
coarse-grained relation scheme.

\paragraph{Comparison with UniRST.}
First, we replace the proposed discourse analysis with
UniRST (Chistova, 2025). We use the
\texttt{tchewik/isanlp\_rst\_v3} parser with the
\texttt{zho.rst.gcdt} relation inventory for Chinese and
\texttt{eng.rst.rstdt} for English. To make its output
compatible with our generator, we map the original
fine-grained relation labels into the six categories used
by our framework. The complete mapping is provided in
Appendix~A, while all other augmentation and training
settings remain unchanged.

Second, we evaluate two ablations. In \textit{w/o Relation},
rhetorical relations are removed from the generation input.
In \textit{Adjacent-only}, each sentence can only select its
immediately preceding sentence as the antecedent, thereby
removing long-distance antecedent search. All other settings
remain unchanged.

\paragraph{Effect of rhetorical relations.}
Removing relation labels reduces performance on all
three evaluated datasets. Under Inp+Con+Aug, w/o Relation obtains
0.240 BLEU and 0.362 ROUGE-L on the Chinese corpus,
compared with 0.276 and 0.400 for the full method. On
Newsroom, its BLEU decreases from 0.780 to 0.618,
while SQuAD EM decreases from 0.731 to 0.570. These
results indicate that sentence order alone is insufficient
to preserve the intended logical connection during
regeneration.

\paragraph{Effect of antecedent selection.}
Adjacent-only retains rhetorical labels but removes
non-adjacent antecedent links. On SQuAD under
Inp+Con+Aug, it obtains 0.573 EM and 0.591 F1,
compared with 0.570/0.615 for w/o Relation and
0.731/0.779 for the full method. It therefore slightly
improves EM over w/o Relation but yields a lower F1
score. The ordering is also inconsistent across
summarization metrics, suggesting that restricting all
antecedents to adjacent sentences does not provide a
stable advantage. Overall, rhetorical labels provide
local guidance, while flexible antecedent selection
captures dependencies beyond adjacent sentences.

\subsection{Generalization Across Architectures}
\label{sec:generalization}

Tables~\ref{tab:summary-2} and~\ref{tab:QA-2} examine whether the effect depends on
LLM fine-tuning using BertSum \citep{liu2019text},
SimCLS \citep{liu2021simcls}, RoBERTa \citep{liu2019roberta}, and ELECTRA \citep{clark2024electra}. On the Chinese corpus,
adding Aug to Inp+Con improves ROUGE-L from
0.492 to 0.512 for BertSum and from 0.493 to 0.511
for SimCLS. On Newsroom, the corresponding
improvements are from 0.851 to 0.867 and from
0.850 to 0.874, respectively. The gains also appear
across BLEU and the other ROUGE metrics.

The same pattern holds for discriminative QA models.
RoBERTa improves from 0.633/0.775 to 0.653/0.795
in EM/F1, while ELECTRA improves from 0.624/0.775
to 0.659/0.798. Under the same architectures and
combined-data settings, MixSumm and Two-Step QA
Augmentation remain below the proposed method.

These results indicate that the benefit is not specific
to Qwen. The augmented data also benefits extractive,
contrastive, and discriminative architectures,
demonstrating its generalization across downstream
model families.

\subsection{Label Transfer Validity}

To directly assess label-transfer reliability, we
manually inspect 100 randomly sampled augmented
instances from the Chinese Abstractive Corpus for
summarization and 100 from SQuAD for QA. For summarization, a transfer
is valid if the selected sentence still captures the main
topic or an important aspect of the augmented document;
96 out of 100 instances satisfy this criterion. For QA,
a transfer is valid if the answer is supported by the
augmented context and reasonably answers the
transformed question; 89 out of 100 instances are
judged valid. These results provide direct evidence that
the alignment-based strategy preserves supervision for
most augmented instances.

To further assess the quality of the augmented documents,
we manually evaluate 50 augmented samples for each of
Chinese, English, Indonesian, and Japanese (200 in total)
on fluency, coherence, and discourse-structure preservation
using a five-point scale, and additionally assess whether the
major rhetorical relations are preserved. Averaged across
the four languages, the scores are 4.11 for fluency, 3.93 for
coherence, and 3.86 for discourse-structure preservation,
with 81.5\% of the evaluated samples preserving the major
relations.

\begin{table}[t]
\centering

\footnotesize
\setlength{\tabcolsep}{3pt}
\renewcommand{\arraystretch}{1.08}

\begin{tabular}{@{}lcccc@{}}
\toprule
\multirow{2}{*}{\textbf{Training Configuration}}
& \multicolumn{2}{c}{\textbf{RoBERTa}}
& \multicolumn{2}{c}{\textbf{ELECTRA}} \\
\cmidrule(lr){2-3}
\cmidrule(lr){4-5}
& \textbf{EM}
& \textbf{F1}
& \textbf{EM}
& \textbf{F1} \\
\midrule

Inp
& 0.611 & 0.748
& 0.618 & 0.749 \\

Con
& 0.620 & 0.762
& 0.602 & 0.750 \\

Aug
& 0.625 & 0.748
& 0.609 & 0.751 \\

Inp+Con
& 0.633 & \underline{0.775}
& 0.624 & \underline{0.775} \\

Inp+Aug
& 0.621 & 0.761
& \underline{0.627} & 0.762 \\

Inp+Con+Aug
& \textbf{0.653} & \textbf{0.795}
& \textbf{0.659} & \textbf{0.798} \\

\midrule

Aug-Two-Step
& 0.383 & 0.589
& 0.358 & 0.565 \\

Inp+Aug-Two-Step
& 0.589 & 0.755
& 0.583 & 0.731 \\

Inp+Con+Aug-Two-Step
& \underline{0.637} & \underline{0.775}
& 0.606 & 0.764 \\

\bottomrule
\end{tabular}
\caption{Performance of RoBERTa and ELECTRA on SQuAD. Inp, Con, and Aug denote the original input, control data, and our augmented data, respectively. Aug-Two-Step denotes the training configuration using Two-Step QA augmentation~\citep{chowdhury2024generative}.}
\label{tab:QA-2}
\end{table}

\section{Conclusion}
\label{sec:conc}

We propose a structure-preserving document-level data
augmentation framework based on discourse structure parsing.
The framework changes document content while preserving sentence
order, antecedent links, and rhetorical constraints,
allowing downstream supervision to remain aligned.

Experiments on three summarization datasets and two
extractive QA datasets show that the augmented data
provides useful and complementary training signals.
Comparisons with LLM-based relabeling and Direct LLM Augmentation show that label validity materially affects downstream performance.
The UniRST comparison and controlled ablations further
show that both rhetorical relations and flexible
antecedent selection contribute to downstream
performance. Improvements with BertSum, SimCLS,
RoBERTa, and ELECTRA indicate that the method is not
limited to LLM fine-tuning.

Overall, the framework is intended to supplement rather
than replace real training data by generating diverse
documents whose supervision remains stable and
reusable.

\bibliography{custom}

\appendix

\section{Mapping UniRST Relations to the Proposed
Relation Categories}
\label{app:unirst-mapping}

UniRST does not define a single universal relation
inventory. Instead, its output labels depend on the
discourse treebank used for training. To make these
outputs compatible with our document generator, we
map the source relation labels predicted by UniRST to
the six categories used in our framework.

In Table~\ref{tab:unirst-mapping}, \textit{coarse relation
families} describe the general semantic functions of the
source relations, such as cause, condition, and
elaboration. They are included only to explain the
mapping criterion. \textit{Source relation labels} refer
to the concrete fine-grained labels processed by our
mapping procedure. The mapping is designed specifically for our augmentation framework and is not intended as an official equivalence between the original discourse inventories.

\begin{table}[t]
\centering
\small
\setlength{\tabcolsep}{3pt}
\renewcommand{\arraystretch}{1.08}

\begin{tabularx}{\columnwidth}{
    @{}
    >{\raggedright\arraybackslash}p{0.28\columnwidth}
    >{\raggedright\arraybackslash}X
    @{}
}
\toprule
\textbf{Target category}
&
\textbf{Source relations mapped to this category}
\\
\midrule

Causality \& Condition
&
\textit{Coarse relation families:}
Cause, Condition, and Enablement.

\textit{Source relation labels:}
\texttt{causal-cause},
\texttt{causal-result},
\texttt{contingency-condition},
\texttt{purpose-attribute}, and
\texttt{purpose-goal}.
\\

\addlinespace[2pt]

Contrast \& Transition
&
\textit{Coarse relation families:}
Comparison, Contrast, and Topic-Change.

\textit{Source relation labels:}
\texttt{adversative-antithesis},
\texttt{adversative-concession}, and
\texttt{adversative-contrast}.
\\

\addlinespace[2pt]

Coordination \& Sequence
&
\textit{Coarse relation families:}
Joint and Temporal.

\textit{Source relation labels:}
\texttt{joint-list},
\texttt{joint-disjunction},
\texttt{joint-sequence}, and
\texttt{joint-other}.
\\

\addlinespace[2pt]

Explanation \& Clarification
&
\textit{Coarse relation families:}
Attribution, Background, Elaboration, Explanation,
Manner-Means, and Topic-Comment.

\textit{Source relation labels:}
\texttt{attribution-positive},
\texttt{attribution-negative},
\texttt{context-background},
\texttt{context-circumstance},
\texttt{elaboration-additional},
\texttt{elaboration-attribute},
\texttt{explanation-evidence},
\texttt{explanation-justify},
\texttt{explanation-motivation},
\texttt{mode-manner},
\texttt{mode-means},
\texttt{topic-question}, and
\texttt{topic-solutionhood}.
\\

\addlinespace[2pt]

Summary \& Generalization
&
\textit{Coarse relation families:}
Evaluation and Summary.

\textit{Source relation labels:}
\texttt{evaluation-comment},
\texttt{restatement-partial}, and
\texttt{restatement-repetition}.
\\

\addlinespace[2pt]

Continuation \& Expansion
&
\textit{Coarse relation families:}
Textual Organization, Same-Unit, Span, and other
structural relations.

\textit{Source relation labels:}
\texttt{organization-heading},
\texttt{organization-phatic},
\texttt{organization-preparation},
\texttt{same-unit}, and other unmatched structural
labels.
\\

\bottomrule
\end{tabularx}

\caption{
Relation label mapping used in our framework, where
fine-grained discourse relation labels predicted by UniRST are grouped
into six target categories. The grouping is designed for document
augmentation purposes and follows semantic correspondence between
relations rather than modifying the original UniRST relation definitions.
}
\label{tab:unirst-mapping}
\end{table}

The mapping follows the dominant semantic function of
each source relation. Causal, conditional, and purposive
relations are grouped together; adversative relations
form the contrast category; multinuclear joint and
temporal relations form the coordination category;
relations that provide background, evidence, attributes,
or manner information form the explanation category;
evaluative and restatement relations form the summary
category. Purely organizational relations are assigned to
Continuation \& Expansion because they do not express
one of the five more specific semantic functions.

\end{document}